\documentclass[conf]{new-aiaa}

\usepackage[utf8]{inputenc}
\usepackage{amsmath}
\usepackage{graphicx}
\usepackage{tikz}
\usepackage{pgfplots}
\usetikzlibrary{arrows.meta,positioning,decorations.pathreplacing}
\usepackage{subcaption}
\usepackage{siunitx}
\usepackage{longtable,tabularx}
\usepackage{multirow}
\usepackage{makecell}
\usepackage{booktabs} 
\usepackage{cleveref}
\usepackage[acronym]{glossaries}
\usepackage{standalone}
\usepackage{xcolor}
\usepackage{hyphenat}
\usepackage{adjustbox}

\usepgfplotslibrary{groupplots,fillbetween}
\pgfplotsset{compat=1.18}
\definecolor{pastelBlueDark}{HTML}{4A90E2}
\definecolor{pastelBlueLight}{RGB}{130,170,210}
\definecolor{warmOrange}{RGB}{200,120,50}
\definecolor{warmOrangeLight}{RGB}{230,175,110}
\definecolor{taxiwayGray}{RGB}{215,215,215}
\definecolor{centerYellow}{RGB}{227,200,104}
\definecolor{aircraftGray}{RGB}{140,140,148}
\definecolor{ellipseBlue}{RGB}{50,100,180}
\definecolor{renderGreen}{RGB}{40,120,80}
\definecolor{pastelRose}{RGB}{190,100,120}
\glsdisablehyper
\newacronym{vor}{VOR}{very high frequency omni-directional range}
\newacronym{imu}{IMU}{inertial measurement unit}
\newacronym{dme}{DME}{distance measuring equipment}
\newacronym{gnss}{GNSS}{global navigation satellite system}
\newacronym{gps}{GPS}{global positioning system}
\newacronym{pit}{PIT}{probability integral transform}
\newacronym{crps}{CRPS}{continuous ranked probability score}
\newacronym{pf}{PF}{particle filter}
\newacronym{ukf}{UKF}{unscented Kalman filter}
\newacronym{pca}{PCA}{principal component analysis}
\newacronym{kuni}{KUNI}{Ohio University airport}
\newacronym{cnn}{CNN}{convolutional neural network}
\newacronym{dal}{DAL}{design assurance level}
\newacronym{ce}{CE}{cross-entropy}
\newacronym{nll}{NLL}{negative log-likelihood}
\newacronym{ece}{ECE}{expected calibration error}
\newacronym{mce}{MCE}{maximum calibration error}
\newacronym{dino}{DINO}{self-distillation with no labels}
\newacronym{cte}{CTE}{cross-track error}
\newacronym{ate}{ATE}{along-track error}

\crefformat{subsubsection}{#2Sec.~#1#3}

\crefname{section}{Section}{Sections}
\crefname{subsection}{Section}{Sections}
\crefname{subsubsection}{Section}{Sections}
\crefname{paragraph}{Section}{Sections}
\Crefname{section}{Section}{Sections}
\Crefname{subsection}{Section}{Sections}
\Crefname{subsubsection}{Section}{Sections}
\Crefname{paragraph}{Section}{Sections}

\crefname{figure}{Fig.}{Figs.}
\Crefname{figure}{Figure}{Figures}

\crefname{table}{Table}{Tables}
\Crefname{table}{Table}{Tables}

\title{Filtering Framework for Aircraft Localization During Taxiing Using Semantic Segmentation}

\author{Eleanor \hspace{-3pt} J. \hspace{-3pt} Prince\footnote{Equal contribution}\footnote{Graduate Student, Department of Mechanical Engineering}, \hspace{-3pt} Olivia \hspace{-3pt} B. \hspace{-3pt} Bruvik\protect\footnotemark[1]\footnote{Graduate Student, Department of Computer Science}, \hspace{-3pt} Marc \hspace{-3pt} R. \hspace{-3pt} Schlichting\footnote{Postdoctoral Researcher, Department of Aeronautics and Astronautics}, \hspace{-3pt} Sydney \hspace{-3pt} M. \hspace{-3pt} Katz\protect\footnotemark[4], \hspace{-3pt} Caroline \hspace{-3pt} E. \hspace{-3pt} Cahilly\protect\footnotemark[3], \hspace{-3pt} Alexandros \hspace{-3pt} Tzikas\protect\footnote{Graduate Student, Department of Aeronautics and Astronautics}, \hspace{-3pt} Arec \hspace{-3pt} L. \hspace{-3pt} Jamgochian\protect\footnote{Research Associate, Department of Aeronautics and Astronautics},  \hspace{-3pt} Robert \hspace{-3pt} J. \hspace{-3pt} Moss\protect\footnotemark[4], \hspace{-3pt} and \hspace{-3pt} Mykel \hspace{-3pt} J. \hspace{-3pt} Kochenderfer\footnote{Associate Professor, Department of Aeronautics and Astronautics, AIAA Associate Fellow}}
\affil{Stanford University, Stanford, CA, 94305, USA}

\newcommand\blfootnote[1]{%
  \begingroup
  \renewcommand\thefootnote{}%
  \footnotetext{#1}%
  \endgroup
}

\begin{document}

\maketitle
\blfootnote{Presented at the AIAA AVIATION Forum 2026, San Diego, CA, 8--12 June 2026,
AIAA Paper 2026-4491. Copyright \textcopyright\ 2026 by Stanford Intelligent Systems
Laboratory. Published by the American Institute of Aeronautics and Astronautics, Inc.,
with permission. This is the authors' accepted manuscript. The version of record is
available at \url{https://doi.org/10.2514/6.2026-4491}.}

\begin{abstract}
Autonomous airport taxiing requires accurate localization on complex airport surfaces. Backup systems that do not depend on satellite navigation are needed to reduce failure risk and improve operational resilience. Vision-based systems have emerged as a promising alternative, but existing vision-based approaches do not yet provide probabilistic absolute position estimates with calibrated uncertainty. We present an end-to-end pipeline that combines an image segmentation network with a Bayesian filtering approach to produce a probabilistic position estimate. Instead of directly using the image segmentations in the filter, we first compute the principal components to compress the high-dimensional segmentation masks into compact representations. We calibrate the probabilistic position estimate via a post-hoc multivariate Student-$t$ distribution transfer. To support training and evaluation, we release the first publicly available dataset for vision-based airport surface localization, consisting of 10,000 static image-mask pairs and 120 trajectories at Ohio University Airport. We evaluate our framework on a suite of simulated trajectories based on tracking accuracy and calibration of the reported probabilistic estimate. Across 60 test trajectories, the filter achieves a median cross-track error of 0.07\,m, a median along-track error of 1.95\,m, and an expected calibration error of 0.045 after post-hoc distributional recalibration.
\end{abstract}

\section{Introduction} \label{sec:introduction}

\begin{figure*}[t]
    \centering
    \begin{adjustbox}{width=\textwidth}
        \input{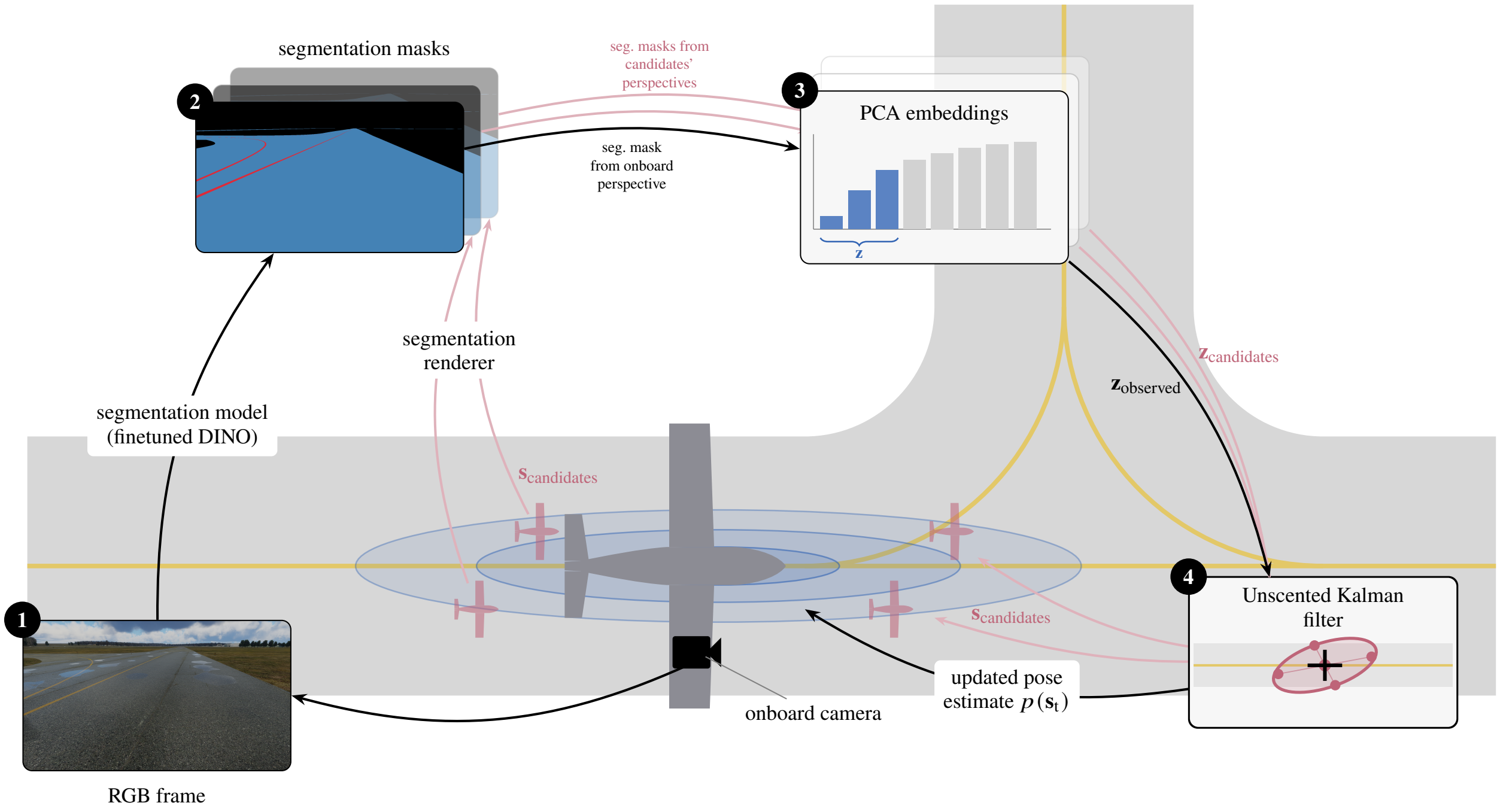}
    \end{adjustbox}
    \caption{High-level overview of the localization pipeline. An onboard RGB image is processed by a segmentation network to produce a segmentation mask, which is projected into a compact \gls{pca} embedding. An unscented Kalman filter compares this embedding against rendered candidate masks from the airport map to maintain a posterior distribution over the aircraft's position.}
    \label{fig:pipeline_overview}
\end{figure*}

While significant progress has been made toward autonomy in airborne navigation and terminal operations, autonomous taxiing remains a distinct and unsolved challenge. Surface operations demand continuous, precise localization within complex taxiway networks, intersections, and dense visual clutter from markings, signage, vehicles, and other aircraft. Current surface localization relies predominantly on \glspl{gnss}~\cite{meier2005benefits}. However, civil aviation's safety guidelines impose many requirements on any navigation source, two of which are particularly relevant here. The first is dissimilar redundancy, where sensing must span independent modalities so that the failure of one system, such as \gls{gnss} jamming or signal loss, does not compromise safety~\cite{faa_ac_1309, rtca_do_178c}. The second is quantified uncertainty, where every position estimate must be accompanied by a measure of confidence, because downstream safety decisions---such as whether to trust an estimate or revert to an alternative source---depend on the reliability of the estimate. \gls{gnss} receivers, for example, report integrity metrics and protection levels alongside their position fixes. These two requirements together therefore motivate a non-\gls{gnss} backup 
that provides position estimates alongside a measure of uncertainty.

Vision-based localization is a natural sensing modality for a backup system, building on the same cues that pilots already use. In current operations, pilots provide visual redundancy by following centerlines, noting hold-short lines, and recognizing pavement markings to confirm their position on the airport surface. Onboard cameras can observe these same surface features, and computer vision algorithms can interpret the resulting imagery to produce position estimates with quantified uncertainty. Recent work in adjacent aviation domains has demonstrated that camera-centric systems can achieve precise state estimation at realistic scales, including probabilistic pose estimation for landing aircraft~\cite{katz2024probabilistic, kawamura2022vision} and vision-based intruder detection during flight~\cite{avoidds}. These results underscore vision as a viable safety-critical modality and motivate its application to surface operations.

\textbf{Vision-based taxiing systems.} Learning-based perception has been explored in prior taxiing work, though existing systems fall short of providing probabilistic absolute localization. TaxiNet~\cite{taxibenchmark} established a foundational \gls{cnn} baseline, estimating cross-track and heading error relative to the runway centerline. However, runways have simple, straight geometry; taxiway networks are far more complex, with intersections, curves, and branching paths. Moreover, TaxiNet's outputs are purely relative---the system approximates the aircraft's offset from the centerline, but cannot determine the aircraft's position on the airport. Liu et al.~\cite{liu2019autotaxiing} moved beyond runways to a full vision-guided taxiing architecture that navigates taxiway networks using air traffic control commands, but vision serves only for local centerline tracking and obstacle detection---the system assumes known global position rather than estimating it from imagery. Gaikwad et al.~\cite{gaikwad2023autotaxi} similarly used \gls{cnn}-based segmentation to detect taxiway lanes and steer with feedback control, but again without estimating the aircraft's absolute position on the airport. No prior vision-based taxiing system provides probabilistic absolute position estimates with quantified uncertainty.

\textbf{Semantic localization.} Outside aviation, the robotics community has shown that semantic segmentation combined with probabilistic localization can achieve long-horizon position tracking using class-level maps---pre-built spatial maps that label regions by semantic category (e.g., road, building, vegetation) rather than raw appearance---allowing localization to remain robust to lighting and weather changes that would confound appearance-based methods~\cite{stenborg2018longterm, herb2021semanticalign, poschmann2017synthsemantic}. In these approaches, a segmentation model classifies each pixel in a camera image into semantic categories, and a filter matches the resulting semantic view against a pre-built map to estimate position. Airport surfaces are a compelling candidate for such techniques: they are dominated by distinctive, structured semantic features---centerlines, hold-short lines, and pavement boundaries---that persist across conditions. Crucially, the geometric complexity of taxiway networks, which poses a challenge for methods like TaxiNet, is precisely what makes approaches based on semantic localization effective: intersections, curves, and branching paths provide rich localization signals that inform the filter's position estimate. However, semantic localization methods have not been applied to aviation surface operations, and the high-precision maps required do not exist for many airports at this time.

\textbf{Uncertainty quantification and calibration.} Adapting semantic localization to airports is only part of the challenge: the resulting probabilistic estimates must also be statistically reliable. A localization system that erroneously reports a confident position estimate can lead to dangerous decisions---for example, trusting a vision-based position that has drifted onto the wrong taxiway. Statistical calibration provides the formal framework for assessing this: a well-calibrated estimator's 90\% confidence region should contain the true position approximately 90\% of the time. Calibration is well-established in meteorology~\cite{gneiting2007scoring} and has been applied to ensemble filters for state estimation~\cite{robert2017lenkpf, robinson2018smoothing}. Despite its importance, no prior vision-based taxiing system has incorporated uncertainty calibration---all existing approaches only produce point estimates~\cite{taxibenchmark, liu2019autotaxiing, gaikwad2023autotaxi}.

Our proposed method provides probabilistic absolute position estimates with calibrated uncertainty for airport taxiing operations. Our contributions are three-fold:
\begin{enumerate}
    \item \textbf{Synthetic data pipeline and public dataset.} We develop an automated pipeline that pairs X-Plane's RGB imagery with ground-truth segmentation masks rendered independently in OpenGL, producing training data at scale without manual annotation. We release the first publicly available dataset for vision-based airport surface localization: $\num{10000}$ static image-mask pairs and $120$ trajectories of $150$ timesteps each.
    \item \textbf{Segmentation with principal components (PCs) used as an embedding.} We fine-tune a self-supervised \gls{dino}~\cite{caron2021dino} segmentation model to extract airport surface features from onboard RGB imagery, and introduce a PC-based embedding that compresses high-dimensional segmentation masks into compact representations.
    \item \textbf{A framework for probabilistic position estimates.} We propose the use of a sample-based filtering method (specifically, a \gls{ukf}) that approximates the posterior position distribution given the PC embeddings of the segmentation masks. We recalibrate the uncertainty prediction of the filter using held-out data. We then evaluate the filtering method out-of-sample along two equally important axes: tracking accuracy and calibration of the reported uncertainty.
\end{enumerate}
The code\footnote{\url{https://github.com/sisl/TaxiLocalizationSemanticFiltering}} and dataset\footnote{\url{https://doi.org/10.25740/zr063zq9238}} are publicly available.

The remainder of this paper is organized as follows. We first review fundamental concepts for our pipeline such as self-supervised visual representations, Bayesian filtering, and calibration. We then present our methods in detail, covering the data generation pipeline, segmentation model, \gls{pca} embedding, filter integration, and recalibration procedure. Finally, we present experimental results and conclude with a discussion of limitations and future directions.

\section{Problem Formulation}
\label{sec:prob_form}

We formulate our problem as a state estimation problem. The aircraft's state at timestep $t$ is its 2D position $\mathbf{s}_t = [x, y]^\top$ on the airport surface in a local coordinate frame. We model the evolution of the state using a simple motion model, and we treat localization as a 2D problem because altitude is known (the aircraft is on the ground) and heading is provided by magnetometers and gyroscopes. These sensors are mechanically independent of satellite signals and therefore do not compromise the dissimilar redundancy argument. Our goal is a non-\gls{gnss} backup for position, not a fully self-contained navigation solution.

At each timestep, three sources of information are available (\cref{fig:pipeline_overview}). First, an onboard camera provides an RGB image $\mathbf{o}_t$. Second, onboard inertial sensors measure the velocity and heading, which are used as the action input $\mathbf{a}_t$ in the motion model for state propagation. Third, a georeferenced airport map $\mathcal{M}$ and known camera parameters $\mathcal{C}_t$ (which include the aircraft's current altitude) together define a rendering function $h(\mathbf{s}, \mathbf{a}, \mathcal{M}, \mathcal{C})$ that generates a synthetic segmentation mask for any aircraft position and action. For any candidate state, we can judge how likely it is to be the true unobserved state by computing the expected observation using the rendering function $h$ evaluated at the candidate state and comparing this expected observation with the obtained observation $\mathbf{o}_t$.

The objective is to estimate the posterior distribution over the aircraft's position at each timestep $t$ given the sequence of observations and actions:
\begin{equation}
\label{eq:main_inference}
    p(\mathbf{s}_t \mid \mathbf{o}_{1:t}, \mathbf{a}_{1:t}).
\end{equation}
The following sections describe how we build such an estimate, starting from high-dimensional RGB images and ending at two-dimensional position estimates via semantic segmentation, compact PC-based embeddings, and a \gls{ukf}.

\section{Preliminaries}

Our pipeline builds on three foundational tools: self-supervised visual representations for semantic segmentation, Bayesian state estimation for probabilistic position tracking, and statistical calibration for assessing the quality of the reported uncertainty. We discuss these in turn. 

\subsection{Self-Supervised Visual Representations} \label{sec:prelim_dino}

Semantic segmentation maps each pixel of an input RGB image to a class label, requiring large labeled datasets where every frame is paired with a ground-truth class map. For specialized domains like airport surfaces, such labeled data does not readily exist and is prohibitively expensive to produce manually at scale. Self-supervised learning addresses this by learning rich visual representations directly from unlabeled imagery as input, producing generalizable feature representations as output---without requiring human annotation. Self-supervised vision transformers (DINO)~\cite{caron2021dino} take augmented views of the same image as input and produce semantically structured feature embeddings as output through self-distillation, which transfer well to downstream tasks. We leverage this transferability to reduce the amount of domain-specific synthetic training data required and improve generalization to unseen airport viewpoints.

\subsection{State Estimation}
\label{sec:bayesian_filtering}
 The inference objective of \cref{eq:main_inference} can be computed recursively on $t$ using a Bayesian update~\cite{thrun2005probabilistic}, where the posterior mean and covariance from the previous timestep are updated using the current observation, the model for the dynamics and the model for the observation. For simple linear dynamics and observation equations and additive Gaussian noise, the update is given in closed form and is known as the Kalman filter~\cite{kalman1960new}.

For general nonlinear models, the update does not have a closed-form solution. Instead, approximate methods such as the \gls{ukf}~\cite{julier2004unscented} propagate a set of deterministically chosen, weighted points (in the case of the \gls{ukf} they are called sigma points) through the nonlinear dynamics to approximate the predicted mean and covariance. These approximations are then used in place of the standard Kalman update to compute the gain and correct the state estimate using the observation $\mathbf{o}_t$ at the current timestep $t$.

\subsection{Calibration of Probabilistic Estimators} \label{sec:calibration_prelim}
 
State estimation algorithms, as described in \cref{sec:bayesian_filtering}, produce not only a point estimate but also a measure of uncertainty. For safety-critical applications, the quality of this uncertainty is as important as the accuracy of the point estimate itself because downstream decisions depend on the predicted confidence being accurate~\cite{thrun2005probabilistic, katz2024probabilistic}.
 
A filter produces a distribution estimate $\mathbb{P}$ that is calibrated if, for any $p \in (0,1)$ and any set $\mathcal{S}$ such that $\mathbb{P}(s \in \mathcal{S}) = p$, the true state falls in $\mathcal{S}$ with empirical frequency close to $p$~\cite{gneiting2007scoring, katz2024probabilistic}. We typically consider sets $\mathcal{S}$ centered at the estimated mean, defined by a threshold on the squared Mahalanobis distance
\begin{equation}
    \delta(\mathbf{s}) = (\mathbf{s} - \hat{\boldsymbol{\mu}}_t)^\top 
    \hat{\boldsymbol{\Sigma}}_t^{-1} (\mathbf{s} - \hat{\boldsymbol{\mu}}_t).
    \label{eq:mahalanobis}
\end{equation}
The distribution of $\delta(\mathbf{s})$ depends on the assumed posterior family: for a Gaussian posterior it follows a $\chi^2$ distribution, while for a multivariate Student-$t$ posterior it follows an $F$ distribution. In both cases, the $p$-confidence region is the set of states whose squared Mahalanobis distance falls below the corresponding $p$-quantile, and the probability of the true state lying within this region is approximately $p$ for a well-calibrated estimator.
 
Calibration is distinct from accuracy: a filter can achieve low mean error while being overconfident (confidence regions that are too small), or conversely, a filter with larger errors can be well-calibrated. 

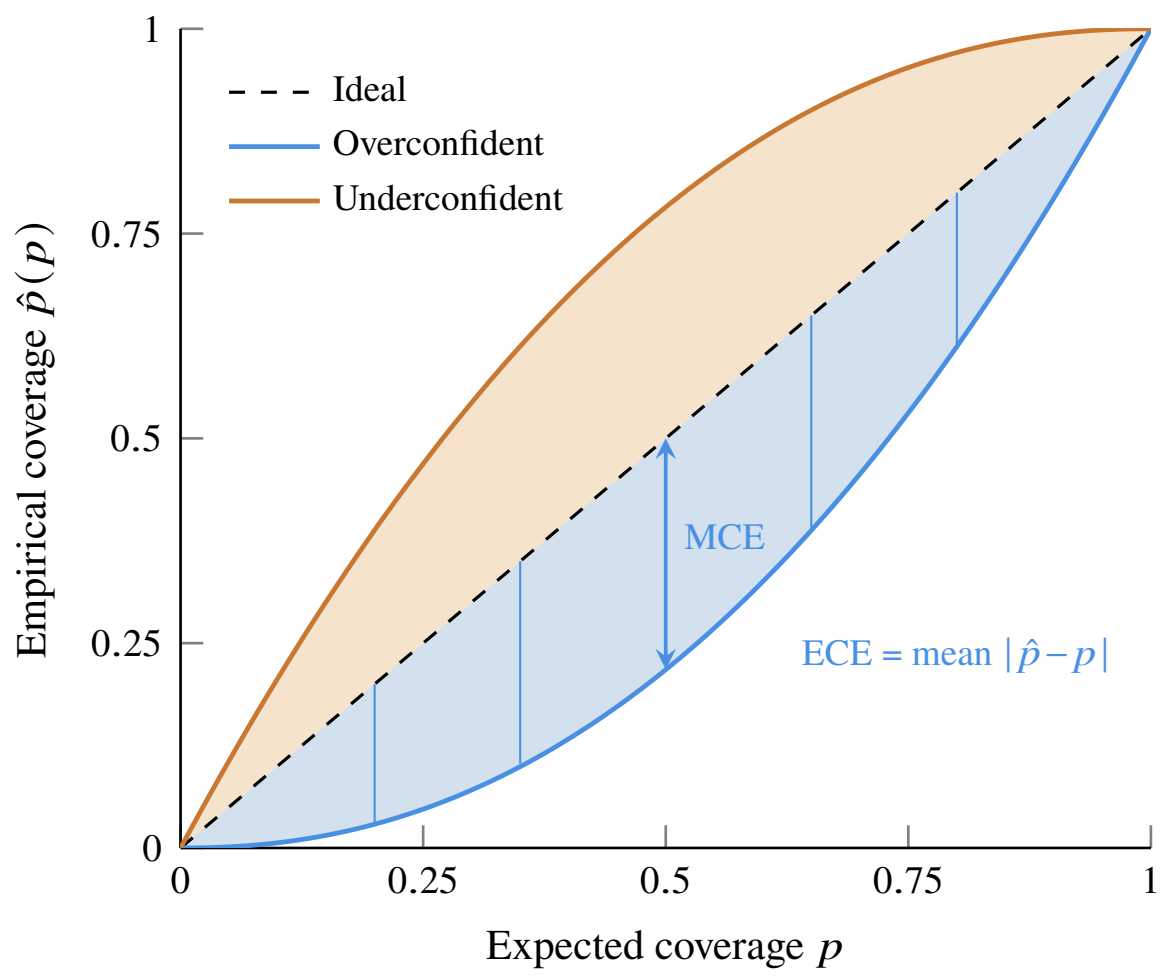
\begin{figure}[t]
    \centering
    \begin{tikzpicture}
\begin{axis}[
  width=8cm, height=7cm,
  xmin=0, xmax=1, ymin=0, ymax=1,
  xtick={0, 0.25, 0.5, 0.75, 1.0},
  ytick={0, 0.25, 0.5, 0.75, 1.0},
  xlabel={Expected coverage $p$},
  ylabel={Empirical coverage $\hat{p}(p)$},
  axis y line=left, axis x line=bottom,
  every outer y axis line/.append style={-},
  every outer x axis line/.append style={-},
  axis line style={line width=0.5pt},
  tick align=inside,
  every tick/.append style={line width=0.5pt},
  label style={font=\footnotesize},
  tick label style={font=\scriptsize},
  legend cell align=left,
  legend style={font=\scriptsize, draw=none, fill=none,
                at={(0.03,0.97)}, anchor=north west},
]

\addplot[black, dashed, line width=0.6pt,
         domain=0:1, samples=2, name path=ideal] {x};
\addlegendentry{Ideal}

\addplot[pastelBlueDark, line width=0.9pt,
         domain=0:1, samples=100, name path=over] {x^2.2};
\addlegendentry{Overconfident}

\addplot[warmOrange, line width=0.9pt,
         domain=0:1, samples=100, name path=under] {1-(1-x)^2.2};
\addlegendentry{Underconfident}

\addplot[pastelBlueLight, opacity=0.35]
  fill between[of=ideal and over];
\addplot[warmOrangeLight, opacity=0.35]
  fill between[of=under and ideal];

\draw[pastelBlueDark, line width=0.4pt]
  (axis cs:0.20,0.030) -- (axis cs:0.20,0.20);
\draw[pastelBlueDark, line width=0.4pt]
  (axis cs:0.35,0.103) -- (axis cs:0.35,0.35);
\draw[pastelBlueDark, line width=0.4pt]
  (axis cs:0.65,0.391) -- (axis cs:0.65,0.65);
\draw[pastelBlueDark, line width=0.4pt]
  (axis cs:0.80,0.614) -- (axis cs:0.80,0.80);

\draw[pastelBlueDark, line width=0.7pt, <->, >=stealth]
  (axis cs:0.50,0.218) -- (axis cs:0.50,0.50);
\node[font=\scriptsize, pastelBlueDark, anchor=west]
  at (axis cs:0.50,0.38) {MCE};

\node[font=\scriptsize, pastelBlueDark, anchor=north, align=center]
  at (axis cs:0.80,0.28) {ECE = mean $|\hat{p}{-}p|$};

\end{axis}
\end{tikzpicture}
    \caption{Conceptual calibration curves. The diagonal (dashed) represents perfect calibration. A curve below the diagonal indicates overconfidence; above indicates underconfidence. The shaded area between a curve and the diagonal visually represents the miscalibration that the \gls{ece} quantifies whereas the maximum distance between the curve and the diagonal is the \gls{mce}.}
    \label{fig:calibration_concept}
\end{figure}

\textbf{Calibration curves.}
Calibration is assessed visually through calibration curves, which plot the empirical coverage $\hat{p}(p)$---the fraction of trials in which the true state falls within the $p$-confidence ellipse---against the nominal expected coverage $p$ (\cref{fig:calibration_concept}). A perfectly calibrated estimator produces a curve that follows the line $\hat{p}(p) = p$. Deviations reveal the nature of miscalibration: a curve below the line $\hat{p}(p)=p$ indicates overconfidence (the estimator's confidence regions are too tight), while a curve above it indicates underconfidence (regions are too conservative).
 
\textbf{Scalar calibration metrics.}
To summarize calibration quality, we use two metrics. The \gls{ece} measures the mean absolute deviation between empirical and nominal coverage:
\begin{equation}
    \text{ECE} = \int_0^1  \left| \hat{p}(p) - p \right|~dp.
    \label{eq:ece}
\end{equation}
The \gls{mce} captures the worst-case deviation:
\begin{equation}
    \text{MCE} = \max_{p \in (0,1)} \left| \hat{p}(p) - p \right|.
    \label{eq:mce}
\end{equation}
Together, these metrics provide a quantitative basis for evaluating the effect of recalibration, as described in \cref{sec:recalibration}.

\section{Methods}

\subsection{Synthetic Data Generation Pipeline} \label{sec:data_generation}

\textbf{Motivation.} Training a segmentation model for airport surfaces requires paired data: an RGB image and a corresponding ground-truth segmentation mask for each viewpoint. No such dataset exists. The flight simulator X-Plane provides photorealistic rendering of airport environments, but it does not expose segmentation-level ground truth. Instead, the airport surface geometry is stored in a proprietary, sparsely documented file format. Our pipeline bridges this gap by independently extracting the airport geometry from X-Plane's data files and rendering ground-truth segmentation masks through a separate OpenGL-based renderer, while using X-Plane for photorealistic RGB imagery and the physics simulation.

\textbf{Geometry extraction.} X-Plane stores airport layouts in text files that describe surface features (taxiway and runway polygons) and line features (centerlines, hold-short markings, edge lines) along with metadata specifying each feature's type and rendering style. However, many geometric properties---such as the widths of different line types---are not documented and are hardcoded in X-Plane's closed-source rendering engine. We reverse-engineered these hardcoded properties and validated the results qualitatively by confirming that the generated segmentation masks align with X-Plane's RGB rendering. Because the geometry files specify only feature boundaries, the parsed polygons and line regions must be triangulated into 3D surface meshes for rendering; we use Gmsh~\cite{geuzaine2009gmsh} for this triangulation step.

\textbf{Ground-truth mask rendering.} An independent renderer built on OpenGL (via \texttt{moderngl}) takes the extracted 3D meshes, camera and heading parameters $\mathcal{C}_t$, and a given aircraft state to produce a dense segmentation mask with three semantic classes: taxiway centerlines, hold-short lines, and pavement (\cref{fig:line_differentiation}). This is the function $h$ as defined in \cref{sec:prob_form}. These classes represent common surface features found at any airport. Taxiway centerlines provide strong cues for cross-track localization; hold-short lines aid along-track localization precisely where it matters most---before entering a potentially active runway; and pavement boundaries constrain the estimate to safe taxiing areas. Because this renderer operates entirely on the extracted geometry, it is independent of X-Plane's rendering engine and can generate masks for arbitrary aircraft states without running the simulator.

\textbf{RGB imagery.} For each aircraft state, X-Plane renders the corresponding photorealistic RGB image from the same camera pose, with full control over environmental conditions (time of day, weather, lighting) through XPlaneConnectX \cite{XPlaneConnectX}. Together with the OpenGL mask renderer, this produces aligned (RGB image, segmentation mask) pairs for any aircraft state on the airport surface.

\textbf{Dataset.} We produce data in two forms. A static dataset consists of image-mask pairs sampled from viewpoints across the airport surface, used for segmentation model training and for fitting the PCA basis. A trajectory dataset consists of sequential frames along taxiway routes, where a proportional controller commands throttle based on velocity error and rudder based on heading error and cross-track error. The trajectory data is used for our filter evaluation and calibration analysis.

\subsection{Segmentation Model} \label{sec:segmentation}

\textbf{Structure.} The first stage of the pipeline converts each RGB image $\mathbf{o}_t$ into a structured semantic segmentation mask $\bar{\mathbf{o}}_t$ that identifies airport surface features. We use a pre-trained DINO model with a ResNet-50 backbone as a frozen feature encoder~\cite{caron2021dino}, followed by a series of learned decoder blocks that progressively upsample the encoded feature maps back to the original image resolution. The output is a three-channel mask at full $1080 \times 1920$ resolution, with one channel per semantic class (taxiway centerlines, hold-short lines, and pavement).

\textbf{DINO model.} The design choice of a frozen DINO backbone with only decoder weights updated is motivated by the properties of self-supervised pre-training discussed in \cref{sec:prelim_dino}. The DINO features already encode rich spatial and semantic structure transferable to the airport surface domain, and fine-tuning only the decoder reduces the risk of overfitting on our synthetic training set. The structured, visually consistent nature of airport surfaces---dominated by geometric patterns such as centerlines, hold-short markings, and pavement boundaries---makes them particularly well-suited to this transfer learning approach. The decoder is trained on the static dataset described in \cref{sec:data_generation} using a deterministic 90/10 train/test split.

\textbf{Loss function.} The loss function is a per-channel multilabel Dice loss, computed independently for each of the three semantic classes and combined via inverse-frequency class weighting. At each training step, per-class pixel counts are computed from the batch and each class loss is assigned a weight inversely proportional to its frequency, normalized so the weights sum to one. This causes the optimizer to attend more strongly to rare, fine features (e.g., centerlines and hold-short lines) relative to the dominant pavement class, which would otherwise drive the loss. The final loss is the weighted dot product of the three per-channel Dice losses. Training hyperparameters are summarized in \cref{tab:hyperparameters}.

\subsection{PCA-based Observation Embedding}

The rendered segmentation masks are high-dimensional: $C \times H \times W = 3 \times 1080 \times 1920 \approx 6 \times 10^6$ values per mask. 
The computational cost of the filtering step scales cubically with the 
observation dimension, because probabilistic filtering requires inverting 
an $m \times m$ observation covariance, where $m$ is the dimension of 
the observation vector.
To remain computationally tractable, the UKF therefore requires a compact, semantically meaningful, representation of the observation.

A natural first attempt is to downsample the masks directly; however, aggressive downsampling is unattractive because the most informative structures, e.g., centerlines and hold-short lines, occupy a narrow set of pixels and are easily erased by uniform downsampling. 
We instead assume that airport surface masks have strong spatial structure and lie near a low-dimensional linear subspace. 
This structure lends itself well to the dimensionality-reducing principal component analysis (\gls{pca})~\cite{greenacre2022principal}. 
PCA finds an orthogonal basis whose directions are ordered by decreasing 
explained variance, so that projecting onto the first $k$ directions 
retains as much variance as possible in $k$ dimensions.
Projecting onto these directions reduces a flattened mask to a $k$-dimensional vector. 
We show in \cref{pca_embedding_quality} that small $k$ retains most of the variance. 

We compute the principal components offline on $N = 9000$ segmentation masks $\bar{\mathbf{o}} \in \{0,1\}^{C \times H \times W}$. 
The computation involves three separate steps. First, we spatially downsample each mask by a factor $d$, producing a lower-dimensional segmentation mask $\bar{\mathbf{o}}' \in \mathbb{R}^{C \times H_S \times W_S}$ that still contains detailed information about finer features with $H_S = \lfloor H/d \rfloor$ and $W_S = \lfloor W/d \rfloor$. 
Second, we flatten each pooled mask into a vector $\mathbf{x}^{(i)} \in \mathbb{R}^D$ with $D = C \cdot H_S \cdot W_S$ and compute the empirical mean $\boldsymbol{\mu} = \tfrac{1}{N}\sum_{i=1}^{N} \mathbf{x}^{(i)}$. 
Stacking all $N$ downsampled segmentation masks row-wise and subtracting the mean, we get a centered data matrix $\mathbf{X} \in \mathbb{R}^{N \times D}$. Third, we fit a rank-$k$ truncated PCA basis offline using singular value decomposition on the centered data matrix $\mathbf{X}$, retaining the top-$k$ right singular vectors $\mathbf{V}_k \in \mathbb{R}^{D \times k}$.
We store $(\boldsymbol{\mu}, \mathbf{V}_k, d)$ for reuse at inference. 
Choices of $k$ and $d$, together with their empirical effect on tracking accuracy and reconstruction quality, are reported in \cref{pca_embedding_quality}.

At each filter step, we apply the same embedding to both the observed mask and the rendered mask of every candidate (a hypothesized aircraft position drawn from the prior distribution $p(\mathbf{s}_t)$).
Each mask is downsampled, flattened (this is denoted $\mathbf{x}$ below), mean-centered, and projected onto the $k$-dimensional basis,
\[
\mathbf{z} = \mathbf{V}_k^{\top} (\mathbf{x} - \boldsymbol{\mu}) \in \mathbb{R}^{k},
\]
yielding the $k$-dimensional observation $\mathbf{z}$ the filter operates on.

\subsection{Filtering} \label{sec:filter_integration}
 
\subsubsection{Dynamics and Observation Models} \label{sec:dynamics}

We instantiate the general Bayesian filtering framework from \cref{sec:bayesian_filtering} with concrete dynamics and observation models specific to aircraft surface operations.

\textbf{Dynamics model.}
The aircraft state evolves according to a linear motion model,
\begin{equation} \label{motion_model}
    \mathbf{s}_{t+1} = f(\mathbf{s}_t, \mathbf{a}_t) + \boldsymbol{\epsilon}_t,
    \qquad \boldsymbol{\epsilon}_t \sim \mathcal{N}(\mathbf{0},\mathbf{Q}),
\end{equation}
where
\begin{equation}
    f(\mathbf{s}_t, \mathbf{a}_t) = \mathbf{s}_t + [v_x, v_y]^\top \Delta t ,
\end{equation}
the action $\mathbf{a}_t = [v_x, v_y, \Delta t]^\top$ is derived from onboard accelerometer and gyroscope measurements, and $f$ denotes the state transition function. Isotropic process noise $\mathbf{Q} = \sigma^2_{\text{proc}} \mathbf{I}$ captures mismatch between the assumed motion model and the true aircraft dynamics. Importantly, the dynamics are linear; all nonlinearity in the pipeline enters through the observation model, which is discussed below.

\textbf{Observation model.}
For a candidate state $\mathbf{s}$ and known action $\mathbf{a}_t$, the predicted observation is not a direct sensor reading but the output of a multi-step rendering-and-embedding pipeline. First, the OpenGL renderer produces a synthetic segmentation mask $\bar{\mathbf{o}}_{\text{synth}} = h(\mathbf{s}, \mathbf{a}_t, \mathcal{M}, \mathcal{C}_t)$ for the candidate position, using the georeferenced airport map $\mathcal{M}$ and camera parameters $\mathcal{C}_t$. Second, the mask is spatially downsampled by factor $d$. Third, it is flattened yielding $\mathbf{x}_\text{synth}(\mathbf{s})$, mean-centered, and projected into the PCA subspace to yield a compact embedding
\begin{equation}
    \mathbf{z}_{\text{render}}(\mathbf{s}) = \mathbf{V}_k^\top(\mathbf{x}_{\text{synth}}(\mathbf{s}) - \boldsymbol{\mu}).
\end{equation}
The camera's RGB image undergoes a similar pipeline---segmentation network, downsample, flatten, project---to produce the corresponding observed embedding $\mathbf{z}_{\text{obs}}$. Both embeddings therefore live in the same $k$-dimensional PCA subspace, enabling direct comparison within a principled probabilistic framework.
 
\subsubsection{Unscented Kalman Filter}

We implement a standard UKF for the two-dimensional position state and the PCA-embedded observation model defined above.
The UKF maintains a Gaussian posterior belief over the aircraft position,
\[
    \mathbf{s}_t \sim \mathcal{N}(\boldsymbol{\mu}_t,\boldsymbol{\Sigma}_t),
    \qquad \mathbf{s}_t \in \mathbb{R}^2 .
\]
At each timestep, the filter performs two unscented transforms: one through the dynamics model and one through the observation model.

At time $t$, using the posterior distribution for $\mathbf{s}_{t-1} \sim \mathcal{N}(\boldsymbol{\mu}_{t-1}, \boldsymbol{\Sigma}_{t-1})$ from the previous timestep, we generate $2n+1$ sigma 
points with $n=2$, labeled $\mathbf{s}_{t-1}^{(i)}$. We weight the mean sigma 
point as $w_1 = \lambda/(n+\lambda)$ and the remaining points by 
$w_i = 1/(2(n+\lambda))$ for $i = 2, \dots, 2n+1$, with a default choice 
of $\lambda = 2$.

In the prediction step, each sigma point $i$ is propagated through the dynamics,
\[
    \mathbf{s}_{t}^{(i)} = f(\mathbf{s}_{t-1}^{(i)}, \mathbf{a}_t).
\]
The predicted mean and covariance are obtained by weighted recombination of the propagated sigma points. They define the predict-step belief.

In the update step, we draw a new sigma-point set from the predict-step belief and pass each point through the observation model described above, rendering $2n+1=5$ predicted observations in $\mathbb{R}^k$.

The standard UKF update is then performed in the $k$-dimensional PCA space, following the formulation of~\cite{kochenderfer2022algorithms}. The sigma points propagated through the observation model yield the predicted observation mean and innovation covariance, from which the Kalman gain is computed and used to correct the predicted state mean and covariance. The true observation made at time $t$ (after being mapped to the $k$-dimensional PCA space) is also used in this update. The observation noise covariance $\mathbf{R} = \sigma^2_{\text{obs}}\,\mathbf{I}$ is isotropic, with $\sigma^2_{\text{obs}}$ selected via Bayesian optimization as described in \cref{sec:hyperparams}. The output of the UKF at time $t$ is $(\boldsymbol{\mu}_t, \boldsymbol{\Sigma}_t)$. 

\subsubsection{Hyperparameter Selection} \label{sec:hyperparams}
We jointly select the following hyperparameters via Bayesian optimization 
using the Optuna framework~\cite{optuna_2019}: the number of principal 
components, the observation and process noise scales, the mask downsampling 
factor, the mask pooling strategy, and whether PCA coefficients are whitened.
We minimize a scalar score that balances tracking accuracy against the calibration of the filter:
\[
\mathcal{L} = \text{position error} + 10 \cdot \text{calibration error},
\]
where the position error is the mean Euclidean distance between the posterior mean and ground-truth aircraft position, and
\[
\text{calibration error} = \sum_{i=1}^{10} |p_i - \gamma_i|
\]
is the summed gap between the empirical coverage $\hat{p}(\gamma_i)$ 
(defined in \cref{sec:calibration_prelim}) and the coverage levels 
$\gamma_i = (10i-1)/100$. The calibration error weight of $10$ was selected empirically from position error analysis and the expected calibration error range.
We optimize over \num{20} trajectories from the training set, discarding the 
first \num{50} timesteps of each trajectory as burn-in, as the filter requires 
time to incorporate enough observations to form a meaningful posterior. 
Both the position error and calibration error are therefore computed over 
the post-burn-in timesteps.

\subsubsection{Recalibration} \label{sec:recalibration}

The filtering pipeline involves several approximations---PCA dimensionality reduction and a Gaussian assumption for the posterior position distribution---that influence calibration. Post-hoc recalibration corrects the uncertainty estimate without retraining the filter.

\textbf{Distributional transfer.} We transfer the filter's output to a multivariate Student-$t$ distribution, which can capture heavier tails that are often observed in the empirical distribution of the state. Concretely, at each timestep $\tau$, we retain the filter's predicted mean $\hat{\boldsymbol{\mu}}_\tau$ and covariance $\hat{\boldsymbol{\Sigma}}_\tau$, but model the posterior distribution as
\begin{equation}
    \mathbf{s}_\tau \sim \mathrm{Student}\text{-}t\left(\hat{\boldsymbol{\mu}}_\tau, \kappa\hat{\boldsymbol{\Sigma}}_\tau, \nu\right),
      \label{eq:recal_student_t}
\end{equation}
where $\nu > 0$ is the degrees-of-freedom parameter controlling tail heaviness and $\kappa > 0$ is a scalar covariance scaling factor. As $\nu \to \infty$ with $\kappa = 1$, the Student-$t$ reduces to the original Gaussian.

\textbf{Recalibration objective.} The parameters $(\nu, \kappa)$ are jointly optimized to minimize the \gls{ece} via
  \begin{equation}
    (\nu^*, \kappa^*) = \underset{\nu > 0, \kappa > 0}{\mathrm{arg\,min}}\ \mathrm{ECE}(\nu, \kappa)
    \label{eq:recal_opt}
  \end{equation}
where the \gls{ece} is evaluated by pooling predictions across the \num{60} 
training trajectories over all timesteps after the filter burn-in period. 
Under the Student-$t$ model, the quantile of each sample is computed via 
the $F$-distribution CDF applied to its squared Mahalanobis distance. We 
use the Nelder-Mead method~\cite{Nelder1965} to optimize in the log-space 
$(\log\nu, \log\kappa)$, which enforces positivity of both parameters.

\section{Experiments and Results}

\subsection{Experimental Setup} \label{sec:experimental_setup}

We evaluate the full pipeline at \gls{kuni}, a general aviation airport whose taxiway geometry is publicly available, enabling the synthetic data pipeline described in \cref{sec:data_generation}. The taxiway network includes straights, curves, and intersections, providing geometric diversity representative of typical airport layouts. Using the pipeline described in \cref{sec:data_generation}, we collect a static dataset of \num{10000} image-mask pairs from viewpoints uniformly distributed across airport taxiways, and a trajectory dataset of \num{120} trajectories of \num{150} timesteps each along taxiway centerlines uniformly distributed across the airport, generated with a diversity of paths via a cascaded proportional controller. All data is collected during daytime hours (9:00~AM--3:00~PM). This restriction is a deliberate scope choice that enables out-of-distribution evaluation under other lighting and weather conditions, which is left for future work. We publicly release this dataset as the first publicly available resource for vision-based airport surface localization research. 
The static dataset is split into \num{9000} training and \num{1000} test image-mask pairs for segmentation network training.
Of the \num{120} trajectories, 60 are the training set for \gls{pca} basis fitting, hyperparameter optimization, and for learning recalibration scaling factors, while 60 are held out for final evaluation. All tracking and calibration results reported in subsequent sections use the 60 held-out test trajectories.

\textbf{Configuration.} The simulated aircraft is a Cessna 172 with a forward-facing camera mounted at the right wing strut attachment point, tilted \SI{20}{\degree} downward, with a horizontal field of view of \SI{90}{\degree} and a resolution of $1080 \times 1920$~pixels at \SI{10}{\hertz}; intrinsic and extrinsic parameters are static with respect to the aircraft frame. 
The filter is initialized with a Gaussian distribution centered at the true starting position with covariance $\operatorname{diag}(0.01, 100)$\,m$^2$, where the diagonal elements give the cross- and along-track uncertainty, respectively. We can typically expect the aircraft to be close to the centerline, but we are more unsure about its along-track position. 
The remaining filter hyperparameters are selected jointly through Bayesian optimization as described in \cref{sec:filter_integration} and reported in \cref{tab:filter_hyperparameters}.

\begin{table}[htbp]
    \centering
    \caption{Filter hyperparameters selected via Bayesian optimization (Optuna).}
    \label{tab:filter_hyperparameters}
    \begin{tabular}{@{}lr@{}}
        \toprule
        Parameter & Value \\
        \midrule
        PCA components $k$         & 21 \\
        Downsampling factor $d$    & 20 \\
        Observation noise $\sigma^2_{\text{obs}}$ & 7.236 \\
        Process noise $\sigma^2_{\text{proc}}$          & 0.216 \\
        Whitening                  & \texttt{false} \\
        Pooling strategy    & adaptive average pooling \\
        \bottomrule
    \end{tabular}
\end{table}

\subsection{Segmentation Performance}

We evaluate the DINO-backbone U-Net model on the held-out test set---comprised of \num{1000} images from the deterministic 90/10 train/test split---using per-class and macro-average Intersection over Union (IoU), and selecting checkpoints based on the highest macro-average IoU. Training hyperparameters are summarized in~\cref{tab:hyperparameters}.

\begin{table}[hbt!]
\caption{Training hyperparameters for segmentation network.}
\label{tab:hyperparameters}
\centering
\begin{tabular}{@{}lr@{}}
\toprule
Hyperparameter & Value \\
\midrule
Batch Size & 8 \\
Epochs & 2 \\
Learning Rate & 0.0001 \\
Loss Function & Dice \\
\bottomrule
\end{tabular}
\end{table}

\textbf{Class segmentation.} The results in \cref{tab:segmentation_iou} show a macro-average IoU of 0.794, with substantial variation across classes that reflects the inherent difficulty of each. Pavement achieves near-perfect IoU (0.987): it is the dominant class by pixel count. Taxiway centerlines, a safety-critical class for localization, achieve an IoU of 0.732, which is encouraging given that they occupy a narrow band of pixels and require precise boundary localization. Hold-short lines have the lowest IoU at 0.665, attributable to their relative rarity in the dataset. The inverse-frequency class weighting applied during training partially compensates for this rarity by upweighting hold-short line loss relative to the dominant pavement class.

\begin{table}[hbt!]
\caption{Per-class and mean IoU on the held-out segmentation test set. The arrow pointing upwards indicates that larger values of IoU are better.}
\label{tab:segmentation_iou}
\centering
\begin{tabular}{@{}lr@{}}
\toprule
Class & IoU$\uparrow$ \\
\midrule
Hold-short lines & 0.6645 \\
Taxiway centerlines & 0.7317 \\
Pavement & 0.9870 \\
\midrule
Mean (macro average) & 0.7944 \\
\bottomrule
\end{tabular}
\end{table}

\begin{figure}[hbt!]
    \centering
    \includegraphics[width=335pt]{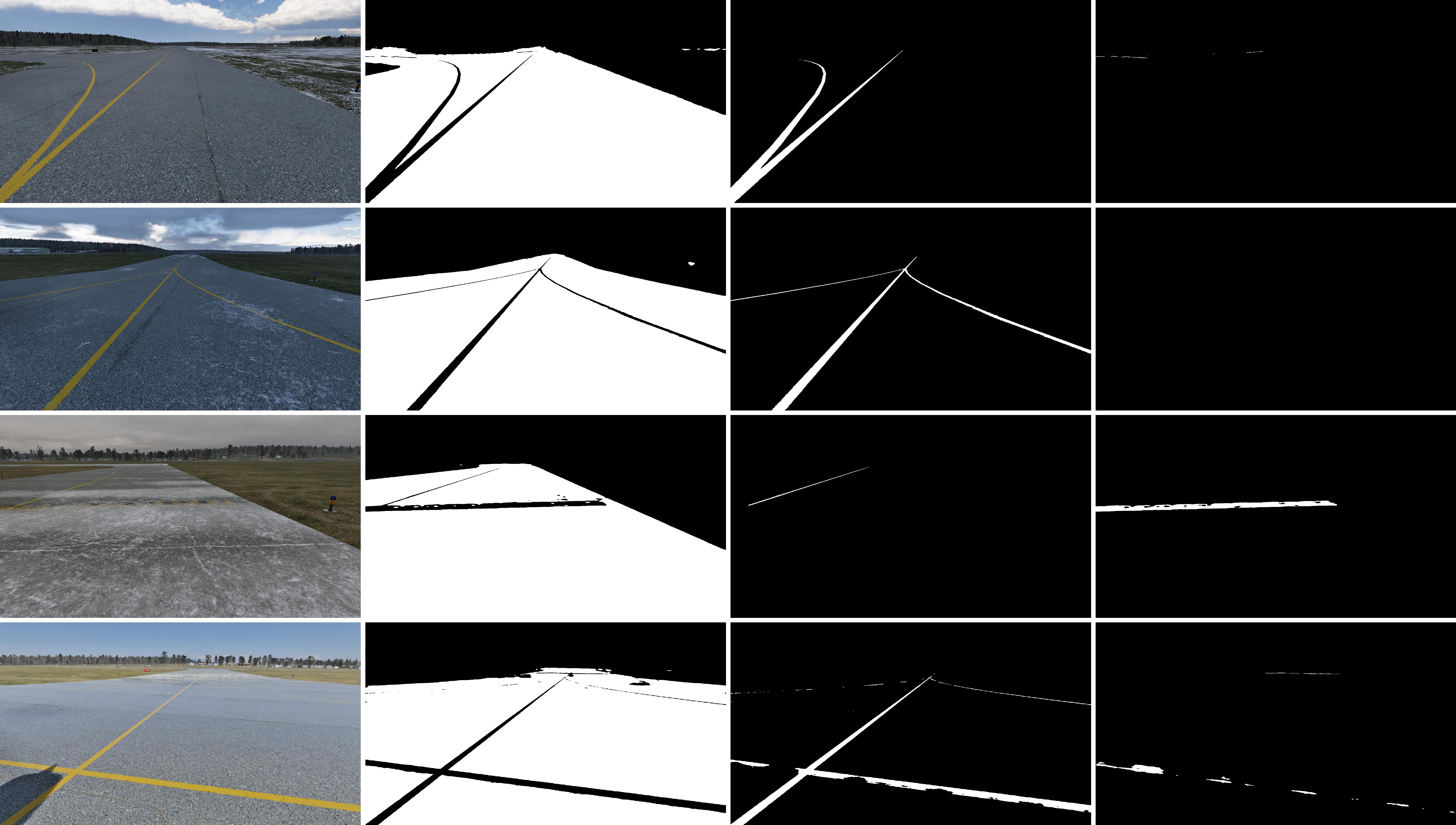}
    \caption{Input RGB frame (leftmost) and predicted segmentation mask channels: hold-short lines (second), taxiway centerlines (third), and pavement (rightmost).}
    \label{fig:line_differentiation}
\end{figure}

\textbf{Line differentiation.} The model successfully distinguishes between the two line types in the vast majority of cases, assigning hold-short lines and taxiway centerlines to separate output channels as shown in \cref{fig:line_differentiation} (Rows 1-3). In rare failure cases---such as the intersection example in Row 4, where the two centerline types meet---the model conflates them, producing overlapping activations in both channels. Nevertheless, such confusion is infrequent, and some residual contribution from these cases may contribute to the lower IoU scores observed for both line types relative to pavement. Extending the pipeline to finer-grained line taxonomies---such as taxiway edge markings and holding position signs---would require additional annotated examples of each class and likely more expressive feature representations to resolve the subtle appearance differences involved.

\subsection{PCA Embedding Quality} \label{pca_embedding_quality}

We compute the \gls{pca} basis offline on $N = 9000$ segmentation masks drawn from the static training set, with an approximately even split between ground-truth and predicted segmentation masks. Each mask is downsampled by a factor of $d = 20$ to a resolution of $54 \times 96$ before fitting the PCA basis. 

\begin{figure}[htbp]
\centering
\includegraphics[width=\textwidth]{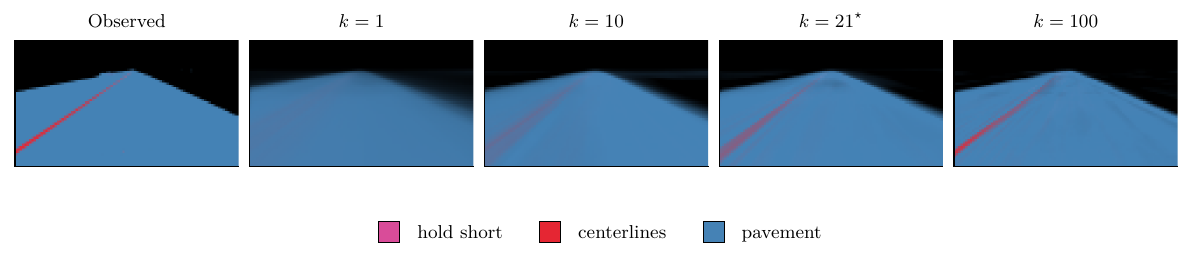}
\caption{PCA reconstruction of a segmentation mask at increasing numbers of principal components $\mathbf{k}$. The class channels are blended into a single image weighted by their probability (legend below). Masks are downsampled to $\mathbf{54 \times 96}$ (downsampling factor of 20) before 
\gls{pca}.
}
\label{fig:pca_reconstruction}
\end{figure}

\Cref{fig:pca_reconstruction} shows reconstructions at increasing values of $k$. At very low $k$, only coarse pavement structure is recovered and fine line features are absent. As $k$ increases, centerlines progressively emerge, with $k = 21$ providing faithful reproduction of all three semantic channels with minimal artifacts. The $k = 21$ selected through the hyperparameter optimization captures 68.0\% of the variance and is validated against tracking accuracy and calibration in \cref{sec:tracking_accuracy}.

\subsection{Tracking Accuracy}
\label{sec:tracking_accuracy}

We evaluate tracking performance using position error (Euclidean distance between the posterior mean and ground truth), decomposed into \gls{cte} and \gls{ate} (the lateral and longitudinal components of the error vector after rotating into the centerline-aligned frame at each timestep). We report mean, median, 75th, and 95th percentile across the $N = 60$ test trajectories, summarized in \cref{tab:multi_traj_error_summary}.

\begin{figure}[h]
\centering
\includegraphics[width=\textwidth]{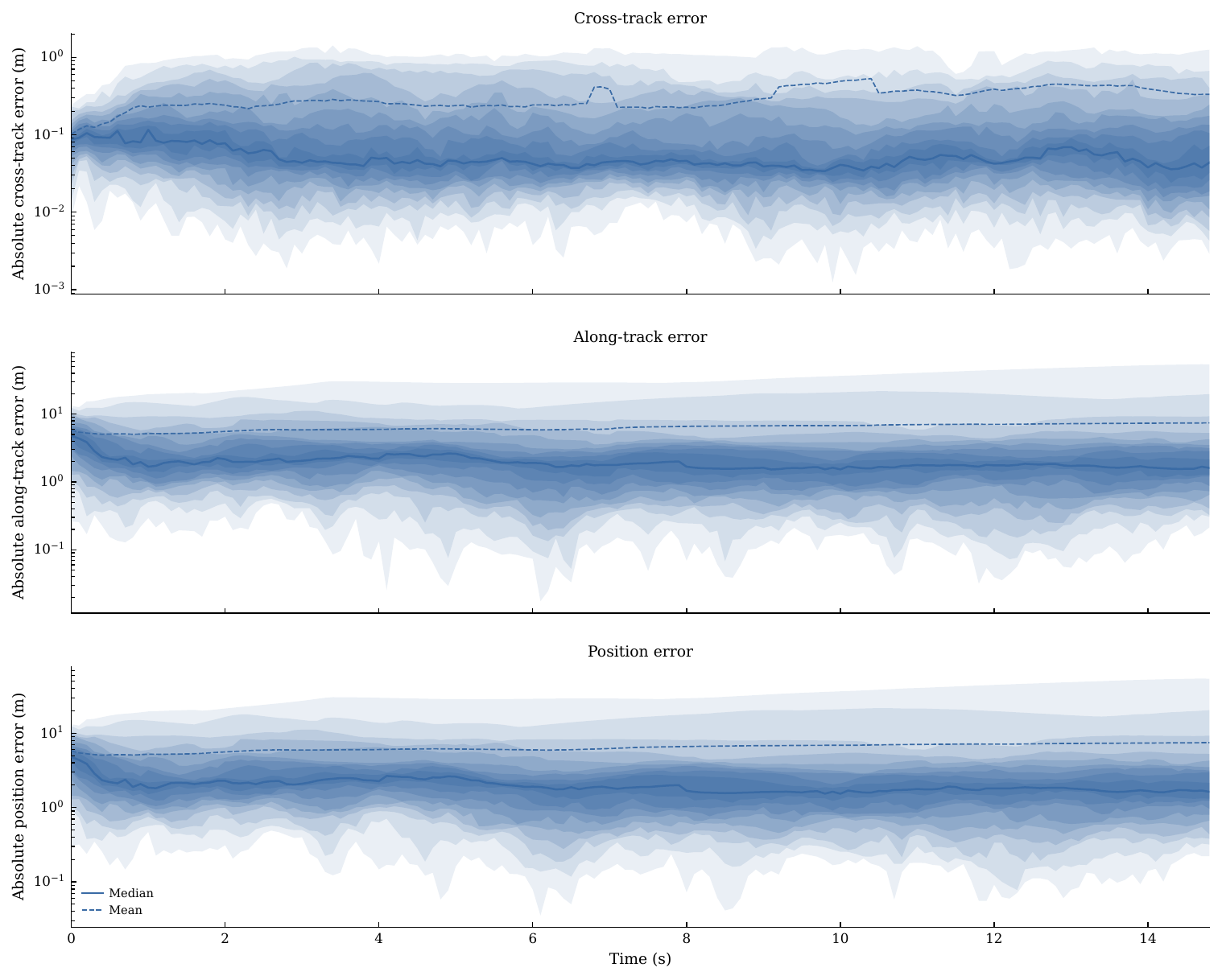}
\caption{Absolute cross-track, absolute along-track, and position error 
    over time across $\mathbf{N = 60}$ test trajectories. Shaded bands show percentile intervals from $5\%$ to $95\%$ in steps of $5\%$; the solid line shows the median, and the dashed 
    line shows the mean.}
\label{fig:accuracy_UKFval}
\end{figure}

\textbf{UKF results.}
Figure~\ref{fig:accuracy_UKFval} shows the time series of \gls{cte}, \gls{ate}, and position error across all \num{60} test trajectories. Mean position error remains approximately flat throughout the trajectory, while the median decreases from $5.79$\,m to $4.77$\,m by the end of the trajectory. The wide gap between mean and median---most visible in the position error panel---reflects a subset of trajectories that fail to converge and persistently inflate the mean. The 75th and 95th percentile bands confirm this: the interquartile range narrows over time, but the upper tail remains large. Averaged over the last $100$ timesteps, the median absolute position error is $1.99$\,m, and among the 51 trajectories that converged (final position error below 10\,m), the median CTE and ATE at convergence were $0.07$\,m and $1.95$\,m respectively.

\begin{figure}[htbp]
\centering
\includegraphics[width=\textwidth]{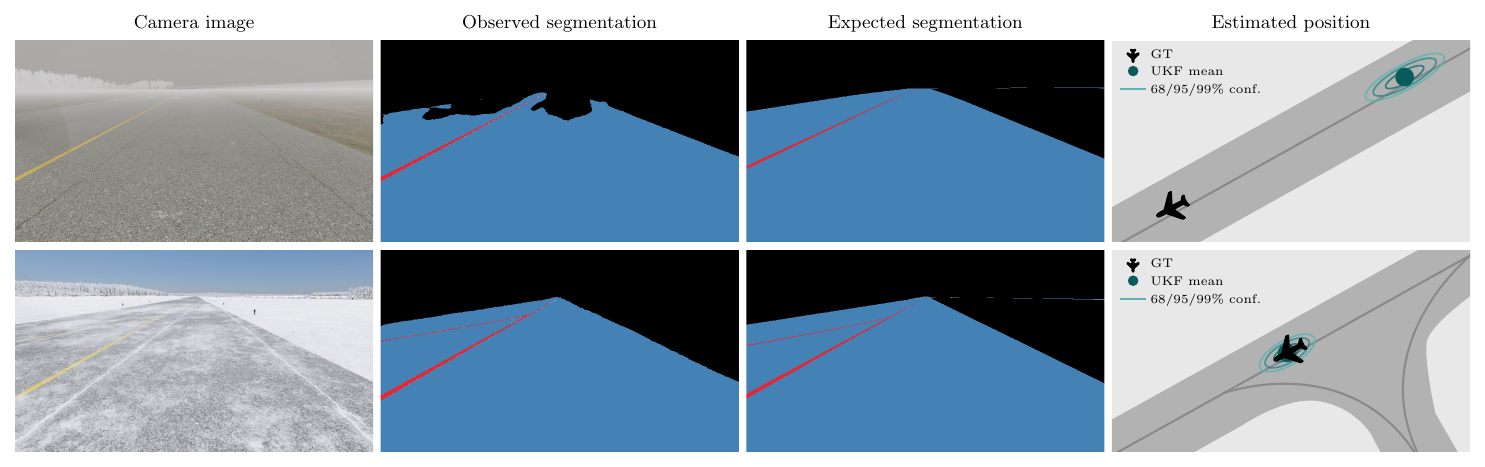}
\caption{Two trajectories at the same timestep (100), contrasting position-estimation failure (top) and success (bottom). In the top panel, poor visibility and a featureless runway produce a noisy CNN segmentation that the rendered expected segmentation does not match precisely. The posterior is displaced from ground truth with overconfident covariance. In the bottom panel, distinctive taxiway markings and higher visibility yield a clean segmentation that aligns with the rendered expectation, and the posterior tracks ground truth closely.}
\label{fig:failure_analysis}
\end{figure}

\textbf{Failure case analysis.}
Manual inspection of non-converging trajectories reveals a common characteristic: they occur predominantly on long straight taxiway segments where centerlines and hold-short markings are absent or outside the camera's field of view. On these segments, the segmentation masks themselves provide minimal localization signal---the observed and rendered masks are nearly identical across a wide range of candidate positions along the taxiway---and the filter cannot correct accumulated drift from the motion model. This failure mode is inherent to the observation model rather than the filter itself: any method that relies on semantic surface features will struggle where those features are absent. These non-converging trajectories inflate both the mean error and the upper percentile bands visible in \cref{fig:accuracy_UKFval}. An example of these failure cases is depicted in \cref{fig:failure_analysis}.

\begin{table}[htbp]
\centering
\caption{Multi-trajectory absolute-error summary across $N=60$ runs. Per metric, the table reports two rows: \texttt{all} aggregates over all $N=60$ runs; \texttt{converged} restricts to the runs whose final-timestep $|\mathrm{pos}|<10$~m. 51 of 60 runs converged (85.0\%). Each row reports the cross-run mean, median, 75th, and 95th percentiles of the per-run last-100-timestep mean. Cross-track and along-track values are computed as $|\,\mathrm{error}\,|$ before aggregation so signed errors do not cancel; position error is already an L2 norm. All values in metres unless otherwise noted. The arrow pointing downwards indicates that smaller values are better.}
\label{tab:multi_traj_error_summary}
\begin{tabular}{l l S[round-mode=places, round-precision=2, table-format=3.2, detect-weight] S[round-mode=places, round-precision=2, table-format=3.2, detect-weight] S[round-mode=places, round-precision=2, table-format=3.2, detect-weight] S[round-mode=places, round-precision=2, table-format=3.2, detect-weight]}
\toprule
{Metric} & {Subset} & {Mean $\downarrow$} & {Median $\downarrow$} & {75th pct} & {95th pct} \\
\midrule
\multirow{2}{*}{Absolute cross-track error} & {all} & 0.3387 & 0.0737 & 0.3300 & 0.8342 \\
 & {converged} & 0.1545 & 0.0506 & 0.1522 & 0.6430 \\
\midrule
\multirow{2}{*}{Absolute along-track error} & {all} & 6.7438 & 1.9462 & 4.2069 & 36.7184 \\
 & {converged} & 1.9898 & 1.3996 & 2.8391 & 4.9765 \\
\midrule
\multirow{2}{*}{Absolute position error} & {all} & 6.8057 & 1.9871 & 4.2081 & 36.7204 \\
 & {converged} & 2.0337 & 1.4544 & 2.8722 & 4.9767 \\
\bottomrule
\end{tabular}
\end{table}

\subsection{Calibration and Recalibration} \label{sec:calibration}

\begin{figure}[hbt!]
\centering
\includegraphics[width=\textwidth]{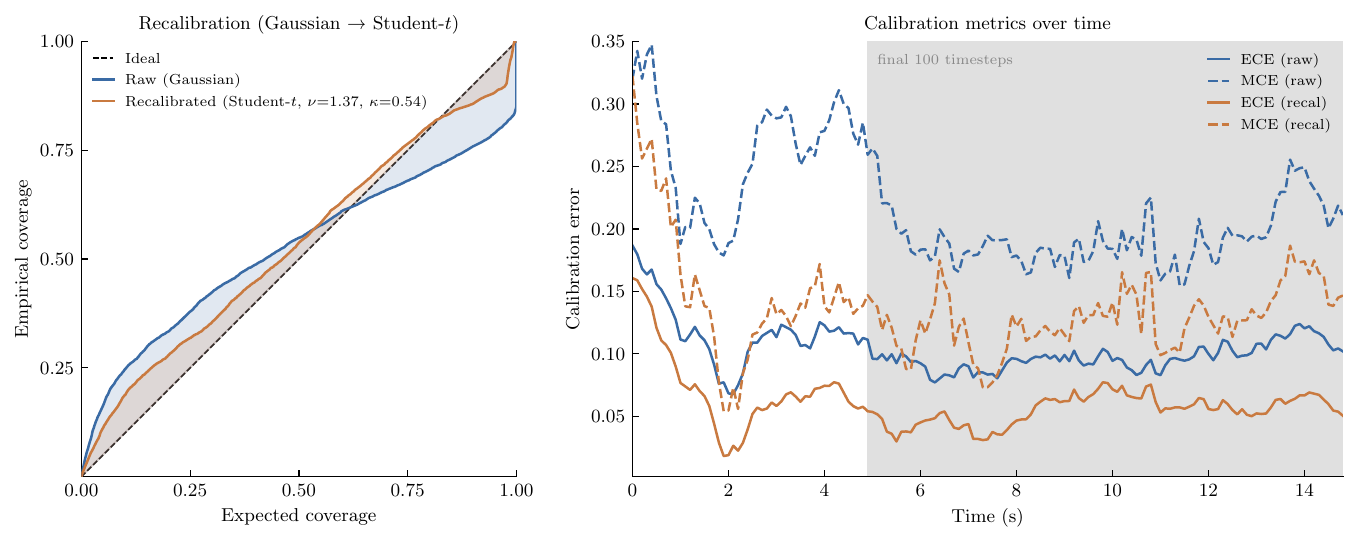}
\caption{Calibration and recalibration results for \gls{ukf}. The recalibration entails the post-hoc transfer from a Gaussian distribution to a Student-$t$ distribution. Left panel shows the calibration curves pooled across the final 100 timesteps to account for burn-in period. The right panel shows \gls{ece} (solid lines) and \gls{mce} (dashed lines) dynamics for the raw \gls{ukf} predictions (blue) and the recalibrated predictions (orange).}
\label{fig:calibration_comparison}
\end{figure}

We evaluate the \gls{ukf}'s uncertainty estimates using the \gls{ece} and \gls{mce} introduced in \cref{sec:calibration_prelim}, computed by pooling predictions across the $N = 60$ test trajectories. Calibration is evaluated over the final $100$ of $150$ timesteps, excluding the initial burn-in period during which the filter has not yet incorporated enough observations to form a meaningful posterior. To protect the integrity of the test dataset, the recalibration parameters $\kappa$ and $\nu$ are computed on the training split of the trajectory dataset. Results are shown in \cref{fig:calibration_comparison} and summarized in \cref{tab:recal_last100}.

\textbf{Raw calibration.} The Gaussian outputs of the \gls{ukf} exhibit an \gls{ece} of $0.094$ and a \gls{mce} of $0.164$ (\cref{tab:recal_last100}). \Cref{fig:calibration_comparison} shows a pronounced s-shaped calibration curve for the raw \gls{ukf} outputs which indicates that the true distribution is more heavy-tailed than the Gaussian distribution output of the \gls{ukf}. 

\textbf{Recalibration.} The heavy-tail nature of the data motivates a distributional transfer for the post-hoc recalibration to a family of distributions that naturally handles heavy tails: the Student-$t$ distribution. 

The optimal recalibration parameters that are found by minimizing \gls{ece} on the training set are $\nu^* = 1.37$ and $\kappa^* = 0.543$ (\cref{sec:recalibration}). The small value of $\nu^*$ indicates that the empirical error distribution has heavier tails than a Gaussian. The scaling factor $\kappa^* = 0.543 < 1$ contracts the covariance, while $\nu < \infty$ produces heavier tails than the Gaussian; together, the scaling and the tail heaviness of the Student-$t$ jointly correct both the bulk and the tails of the distribution.

After recalibration, $\text{ECE}_{\text{cal}} = 0.045$ and $\text{MCE}_{\text{cal}} = 0.093$, representing reductions by factors of $2.09\times$ and $1.77\times$, respectively. The recalibrated curve in the left panel of \cref{fig:calibration_comparison} follows the diagonal substantially more closely across all confidence levels, with the shaded gap between the curve and ideal calibration visibly narrowed. The right panel confirms that recalibration reduces \gls{ece} and \gls{mce} consistently across timesteps throughout the post-burn-in window.

\textbf{Residual miscalibration.} Despite the improvement, slight miscalibration remains after recalibration: an \gls{ece} of $0.045$ and \gls{mce} of $0.093$ indicate that the recalibrated distribution still deviates from ideal by up to approximately 9 percentage points at worst-case coverage levels. We note that even a perfectly specified model evaluated on finite samples would exhibit some statistical deviation from the diagonal, so the residual miscalibration should be interpreted in that context. Analysis of the remaining calibration error is left for future work.

\begin{table}[htbp]
\centering
\caption{Recalibration parameters and metrics (Gaussian $\to$ Student-$t$, pooled, last 100 timesteps).}
\label{tab:recal_last100}
\begin{tabular}{@{}l S[round-mode=places, round-precision=3, table-format=1.3]@{}}
\toprule
{Quantity} & {Value}\\
\midrule
$\nu^*$ & 1.37 \\
$\kappa^*$ & 0.543 \\
$\mathrm{ECE}_{\mathrm{raw}}$ & 0.0942 \\
$\mathrm{ECE}_{\mathrm{cal}}$ & 0.0450 \\
$\mathrm{ECE}_{\mathrm{raw}}/\mathrm{ECE}_{\mathrm{cal}}$ & 2.09 \\
$\mathrm{MCE}_{\mathrm{raw}}$ & 0.1641 \\
$\mathrm{MCE}_{\mathrm{cal}}$ & 0.0926 \\
$\mathrm{MCE}_{\mathrm{raw}}/\mathrm{MCE}_{\mathrm{cal}}$ & 1.77 \\
\bottomrule
\end{tabular}
\end{table}

\section{Discussion}

We presented an end-to-end pipeline that establishes vision-based localization with 
calibrated uncertainty as a viable \gls{gnss} backup for airport surface operations. The pipeline integrates four components: synthetic data generation pairing X-Plane RGB imagery with independently rendered ground-truth masks; a fine-tuned \gls{dino} segmentation network; a \gls{pca} observation embedding; and probabilistic filtering via a \gls{ukf}, followed by post-hoc covariance recalibration. The \gls{pca} embedding is central to tractability, reducing the observation space from roughly six million dimensions to 21---making the \gls{ukf} Kalman gain computation trivial. Alongside the pipeline, we release the first publicly available dataset for vision-based airport surface localization: \num{10000} static image-mask pairs and $120$ trajectories.

Across 60 test trajectories, the filter delivers a median cross-track error of $0.07$\,m and a median along-track error of $1.95$\,m. Failure cases occur predominantly on straight taxiway segments with sparse features, where the observation model provides insufficient signal to correct drift---a limitation inherent to any semantics-based approach. The raw filter output is slightly miscalibrated, with an \gls{ece} of 0.094; post-hoc distributional recalibration via a multivariate Student-$t$ transfer reduces this by a factor of $2.09\times$ to 0.045, with consistent improvement across the post-burn-in window.

Several limitations should temper these findings. Evaluation was confined to synthetic imagery from a single airport, and generalization to real-world imagery and other airports remains to be demonstrated. Heading is assumed known from independent magnetometer and gyro measurements, making the system a backup for position rather than a fully self-contained navigation solution. Recalibration factors are airport- and distribution-specific, and the segmentation model was trained only on daytime imagery (9:00~AM--3:00~PM), leaving open performance under night, adverse weather, seasonal variation, and degraded paint. Finally, isotropic covariance scaling cannot correct anisotropic miscalibration---for example, when cross-track uncertainty is well-calibrated but along-track is not.

Future work follows naturally from these limitations. Most pressing is validation on real onboard imagery across multiple airports and conditions, which would establish whether the synthetic-to-real gap can be closed without retraining. Within the filter itself, heading estimation could be incorporated jointly rather than assumed known, moving the system closer to a fully self-contained navigation solution. Convergence failures on featureless segments call for more robust filter designs, whether through adaptive initialization, online noise tuning, or more expressive architectures that maintain multimodal beliefs when the observation provides insufficient signal to collapse to a single mode. On the safety side, the calibrated uncertainty estimates open a natural path toward integrity monitoring, where the filter flags its own position fixes as unreliable when reported coverage falls below a threshold.

\section*{Acknowledgments}
This work was supported by the NASA University Leadership Initiative (ULI) under grant number 80NSSC24M0069, titled Safe, Scalable, and Seamless SurfNav4UAS. 
Large language models were used to improve the clarity and readability of this manuscript and to assist in the construction of figures. All AI-generated content was reviewed and edited by the authors, who take full responsibility for the final content.

\bibliography{sample}

\end{document}